\documentclass[11pt]{article}

\usepackage[preprint]{acl}

\usepackage{times}
\usepackage{latexsym}
\usepackage{booktabs}
\usepackage[T1]{fontenc}
\usepackage[table]{xcolor}

\usepackage[utf8]{inputenc}

\usepackage{microtype}

\usepackage{inconsolata}

\usepackage{graphicx}

\usepackage{amsmath}

\title{ChartAct: A Benchmark for Dynamic Chart Understanding}

\author{
\textbf{Muye Huang}\textsuperscript{1,2,*},
\textbf{Lin Wu}\textsuperscript{1,2,*},
\textbf{Lingling Zhang}\textsuperscript{1,2,\textdagger},
\textbf{Hang Yan}\textsuperscript{1,2},
\textbf{Zhiyuan Wang}\textsuperscript{1,2}, \\
\textbf{Yumeng Fu}\textsuperscript{1,2},
\textbf{Zesheng Yang}\textsuperscript{1,2},
\textbf{Jun Liu}\textsuperscript{1,2} \\
$^1$School of Computer Science and Technology, Xi'an Jiaotong University \\
$^2$MOE KLNN Lab, Xi'an Jiaotong University \\
\texttt{\{huangmuye, wl19503611685, shihanghanya233, 2444821229\}@stu.xjtu.edu.cn} \\
\texttt{\{yumfuu, youngzsh\}@stu.xjtu.edu.cn, \{zhanglling, liukeen\}@xjtu.edu.cn} \\
$^*$Equal contribution.
\quad
$^\dagger$Corresponding author.
}

\begin{document}
\maketitle
\begin{abstract}
Charts are widely used to present complex data for analysis and decision making. Existing chart understanding benchmarks mainly focus on static charts, but real-world charts are often dynamic and interactive. Key information may only appear after actions such as hovering, clicking, zooming, or dragging. Dynamic chart understanding therefore requires models to identify visible content, choose proper interactions, and reason over changing chart states. To evaluate this ability, we propose ChartAct, an interactive benchmark for dynamic chart understanding. ChartAct collects and filters 673 dynamic charts from 8 real chart websites, covers 7 common chart types, and constructs 1,440 high-quality question-answer samples. Each sample is instantiated in two environments, Dynamic Chart and Dashboard Chart, to evaluate dynamic chart understanding under different contexts. Based on ChartAct, we systematically evaluate 11 advanced multimodal models and GUI agents. Experimental results show that existing models still have clear limitations in dynamic chart understanding. The strongest model, Claude-Opus-4.7, achieves an average success rate of 84.5\%, while most models remain below 60\%. We also conduct detailed failure attribution and case analysis. ChartAct provides a new benchmark for studying chart understanding in real interactive environments. Codes at \url{https://github.com/wulin-wulin/OSWorld_Chart}

\end{abstract}

\begin{figure*}[ht]
    \centering
    \includegraphics[width=1\linewidth]{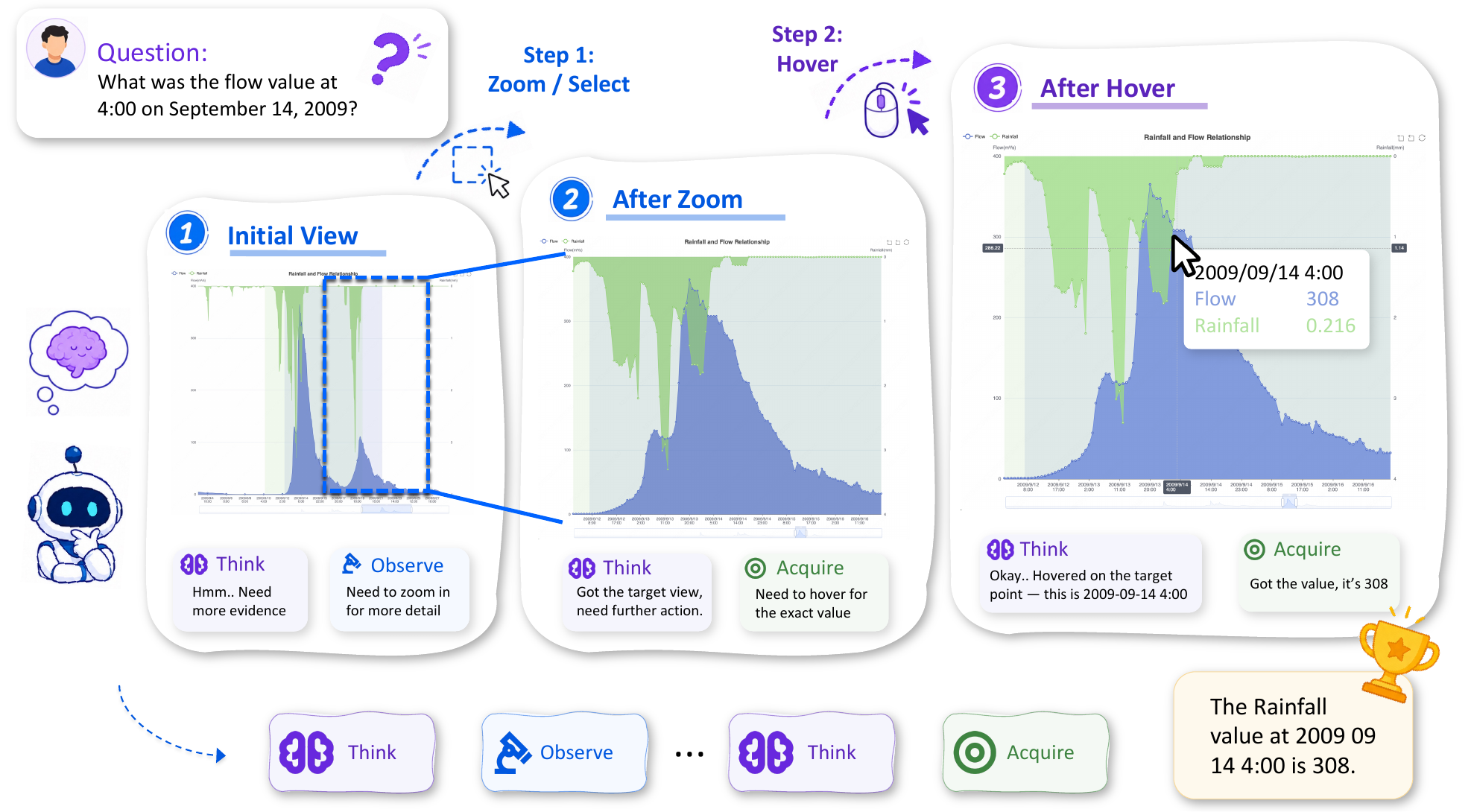}
    \caption{Illustration of dynamic chart understanding in ChartAct. The model starts from the initial chart state, performs actions such as zooming and hovering to reveal hidden evidence, and answers the question based on the newly observed chart state.}
    \label{fig:placeholder}
\end{figure*}

\section{Introduction}
Charts are a widely used form of data visualization, capable of presenting complex data relationships in an intuitive manner. Charts commonly appear in scientific papers, business reports, financial analysis, and public data platforms. Automated chart understanding is an important step toward automated data analysis. This task requires models to understand axes, legends, text, colors, shapes, and data points in charts, and further perform a series of complex inferences. In recent years, extensive chart understanding benchmarks \cite{DBLP:conf/wacv/MethaniGKK20, DBLP:conf/cvpr/KaflePCK18, DBLP:conf/wacv/ChaudhrySGMBJ20, DBLP:conf/iclr/KahouMAKTB18, masry2025chartqaprodiversechallengingbenchmark} have driven substantial progress of MLLMs \cite{Qwen-VL, wang2024qwen2vlenhancingvisionlanguagemodels, masry2024chartgemmavisualinstructiontuningchart, li2024llava, DBLP:journals/corr/abs-2310-11441} in chart understanding.

However, existing chart evaluation benchmarks focus on static charts, where all information is presented through a static image. This setting does not fully match the form of charts in real scenarios. Charts in real scenarios usually have dynamic and interactive properties, which we refer to as dynamic charts. A large number of dynamic charts are embedded in webpages or dashboards, where users need to interact with the interface to obtain the required information. When analyzing charts, users may hover over data points to inspect precise values, click legends to switch data series, or drag time sliders to observe changes across different stages. In these scenarios, key information is often not fully available in the initial chart state. Models need to obtain new information through interface actions and complete reasoning based on continuously changing chart states.

This process imposes higher requirements on chart understanding. First, models need the ability to interact with chart environments. The evidence required by real questions often needs to be obtained progressively through actions, so models must select appropriate interactions according to the question and update the subsequent reasoning process based on new observations. Second, models need to precisely understand and locate visual elements and interactive elements in dynamic charts. Markers, controls, and interface states in dynamic charts jointly determine the currently visible information. If a model cannot accurately locate target elements or understand their functions, it is difficult to perform effective interactions. Existing static chart evaluations mainly focus on visible information in a single image, and thus cannot adequately evaluate interaction decision-making, element localization, and evidence acquisition in dynamic chart environments.

To study this problem, we propose ChartAct, a benchmark for dynamic chart understanding. ChartAct places dynamic charts in interactive environments, allowing models to change chart states through actions, as Figure~\ref{fig:placeholder} shows. Changes in chart states produce new observable information, which provides evidence for the model's subsequent reasoning. This process is close to real dynamic chart analysis. When analyzing dynamic charts, users usually operate the chart repeatedly according to the question, observe the feedback, and update their judgments. ChartAct evaluates the dynamic chart understanding ability of models around this process. To support this evaluation, we need to construct an evaluation environment that preserves both the complexity of real charts and the controllability of interaction processes.

Specifically, we crawl 673 dynamic charts from 8 real websites, and manually filter the collected charts to retain samples with clear data semantics, rich interactive behaviors, and high analytical value. We further embed these dynamic charts into controllable interactive environments, constructing interactive evaluation environments that contain charts, titles, controls, and other components. Based on these environments, we design tasks covering visible-state question answering, interaction-revealed information, cross-state comparison, multi-chart evidence integration, and interactive context understanding. We evaluate current advanced multimodal models and GUI agents \cite{DBLP:conf/acl/NguyenCW0PHLWA025} on ChartAct, and analyze their failure modes in chart localization, action selection, state observation, and evidence integration.

Our main contributions are summarized as follows:

\begin{itemize}
    \item We propose ChartAct, a benchmark for dynamic chart understanding, which evaluates the ability of models to acquire chart evidence through actions and complete reasoning in interactive environments.

    \item We construct an evaluation set based on dynamic charts from real websites. We manually filter high-value chart samples and embed them into controllable simulated dashboards, preserving both real chart complexity and reproducible interactive environments.

    \item We systematically evaluate advanced multimodal models and GUI agents, revealing key limitations of current models in chart localization, interaction action selection, state observation, and cross-state evidence integration.
\end{itemize}

\section{Related Work}

\subsection{Chart Understanding Datasets}

Chart understanding datasets are the basis for evaluating chart perception and reasoning. Early benchmarks such as FigureQA \cite{DBLP:conf/iclr/KahouMAKTB18}, DVQA \cite{DBLP:conf/cvpr/KaflePCK18}, and PlotQA \cite{DBLP:conf/wacv/MethaniGKK20} mainly use synthetic charts and template-based questions. Later datasets, including ChartQA \cite{DBLP:conf/acl/MasryLTJH22, masry2025chartqaprodiversechallengingbenchmark}, RealCQA \cite{DBLP:conf/icdar/AhmedJPSG23}, ChartX \cite{DBLP:journals/corr/abs-2402-12185}, UniChart \cite{DBLP:conf/emnlp/MasryKLHJ23}, expand chart sources and training data. ChartQA introduces real charts and human-written questions, while UniChart provide larger-scale data by integrating existing datasets or generating new samples. Recent benchmarks such as CharXiv \cite{DBLP:journals/corr/abs-2406-18521} and EvoChart \cite{DBLP:conf/aaai/HuangLZW0ZL25} further move chart evaluation toward real world scenario charts. DashboardQA \cite{DBLP:conf/eacl/KarthaMILRMRAPHJ26} focus on dashboards. Most existing datasets still focus on static charts. ChartAct extends this direction to dynamic charts, where models need interaction to acquire chart evidence.

\subsection{MLLMs for Chart Understanding}

MLLMs have become a major approach for chart understanding. Earlier methods such as DePlot  \cite{liu2023deplotoneshotvisuallanguage}, MatCha  \cite{DBLP:conf/acl/0001PKPLJACE23}, UniChart \cite{DBLP:conf/emnlp/MasryKLHJ23}, and ChartReader \cite{DBLP:conf/iccv/ChengDH23} improve chart question answering through chart parsing or chart-specific pretraining. Recent methods, including ChartLlama \cite{DBLP:journals/corr/abs-2311-16483}, ChartAssistant \cite{DBLP:journals/corr/abs-2401-02384}, ChartInstruct \cite{DBLP:journals/corr/abs-2403-09028}, TinyChart \cite{DBLP:journals/corr/abs-2404-16635}, and ChartMoE \cite{xu2025chartmoemixturediverselyaligned}, use chart instruction data and specialized architectures to improve chart perception and reasoning. General MLLMs \cite{chen2023internvl, gemmateam2025gemma3technicalreport, qvq-72b-preview, bai2025qwen25vltechnicalreport, abdin2024phi3technicalreporthighly} also achieve strong results on static chart benchmarks such as ChartQA. These methods mainly answer from a single chart image. ChartAct evaluates models in dynamic chart environments, where models must observe state changes and reason from interactively acquired evidence.

% GUIAgents: A Survey
% OSWORLD: Benchmarking Multimodal Agents for Open-Ended Tasks in Real Computer Environments
% Mobile-Agent-v3: Fundamental Agents for GUI Automation
\subsection{GUI Agents}

GUI agents operate graphical interfaces through visual perception, grounding, and sequential actions. Recent benchmarks and systems \cite{DBLP:conf/eacl/LiuLWXHXZHYW26, DBLP:conf/aaai/YangWTZCJL26, DBLP:conf/acl/ChenCHQFZWLCHYL25, DBLP:conf/iclr/GouWZXCS0025}, such as OSWorld and Mobile-Agent-v3 \cite{DBLP:conf/nips/XieZCLZCHCSLLXZ24, DBLP:journals/corr/abs-2508-15144}, evaluate agents in executable desktop or mobile environments and study their abilities in interface grounding, planning, and environment feedback. These works show strong progress in general GUI automation, but they mainly focus on broad application-level tasks. ChartAct studies a more focused setting, where agents must control the interface to reveal chart evidence and then reason over visual data. This setting connects GUI interaction with chart understanding.

\begin{figure*}[ht]
    \centering
    \includegraphics[width=1\linewidth]{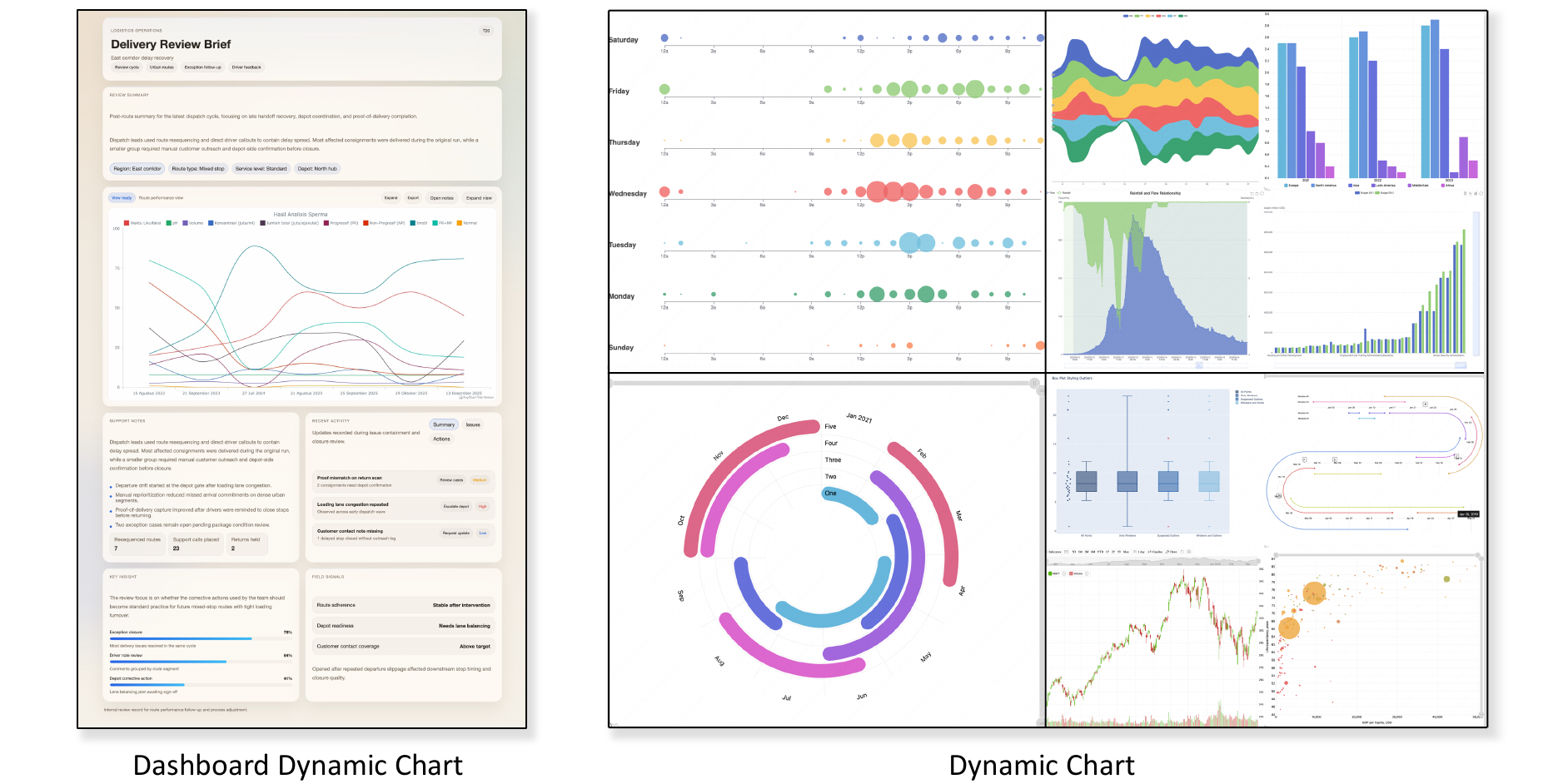}
    \caption{Examples of the two evaluation environments in ChartAct. Dashboard Chart embeds a dynamic chart into a dashboard page, while Dynamic Chart presents the same type of interactive chart in a clean chart environment.}
    \label{fig:overview}
\end{figure*}
\begin{table*}[ht]
\small
\centering
\setlength{\tabcolsep}{9pt}
\renewcommand{\arraystretch}{1.08}
\begin{tabular*}{\textwidth}{@{\extracolsep{\fill}}lrrrrrrrr@{}}
\toprule
\textbf{Source} & \textbf{Line} & \textbf{Bar} & \textbf{Bubble} & \textbf{Pie} & \textbf{Heatmap} & \textbf{Box} & \textbf{Other} & \textbf{Total} \\
\midrule
Apache ECharts & 13 & 13 & 13 & 13 & 13 & 13 & 4 & 82 \\
VChart & 13 & 13 & 13 & 13 & 2 & 3 & 4 & 61 \\
Highcharts & 13 & 18 & 8 & 11 & 7 & 2 & 13 & 72 \\
AnyChart & 16 & 18 & 15 & 15 & 9 & 20 & 8 & 101 \\
Chart.js & 40 & 10 & 6 & 8 & 0 & 0 & 4 & 68 \\
amCharts & 16 & 18 & 11 & 14 & 8 & 0 & 18 & 85 \\
Plotly & 17 & 19 & 17 & 11 & 12 & 19 & 14 & 109 \\
AntV G2 & 19 & 19 & 19 & 19 & 9 & 10 & 0 & 95 \\
\midrule
\textbf{Total} & \textbf{147} & \textbf{128} & \textbf{102} & \textbf{104} & \textbf{60} & \textbf{67} & \textbf{65} & \textbf{673} \\
\bottomrule
\end{tabular*}
\caption{Distribution of collected interactive charts across sources and chart types.}
\label{tab:chart-source-type}
\end{table*}
\section{ChartAct Benchmark}
We propose ChartAct, an interactive evaluation benchmark for dynamic chart understanding. ChartAct centers on real dynamic charts, places them in controllable interactive environments, and requires models to answer chart questions through actions, observations, and reasoning. Each sample consists of a dynamic chart, an interactive environment, a question, and a ground-truth answer. A model starts from the initial chart state, selects interaction actions according to the question, and observes new visible information after the chart state changes. It provides evidence for subsequent reasoning. Based on this process, ChartAct evaluates models' interaction decision-making, state observation, and evidence integration abilities in dynamic charts.

\subsection{Data Construction}

\paragraph{Chart collection.}
We collect dynamic charts with interactive functions from real websites. To ensure that the charts are suitable for interactive evaluation, we manually filter the candidate charts. The filtering process mainly removes two types of samples. The first type contains charts with weak interactivity. These charts may include animations or visual changes, but the changes mainly occur at the visual presentation level. Their chart content does not change effectively through interaction, and the interaction process does not provide new data evidence. The second type contains charts that are unsuitable for interactive tasks. These charts may change continuously and lack a stable observable state, making it difficult for models to observe, operate, and answer under a clear state. After filtering, we retain 673 interactive charts from 8 real websites. These charts cover 7 types, including line charts, bar charts, scatter/bubble charts, pie/donut charts, heatmaps, box/candlestick charts, and other charts. The chart types are kept relatively balanced within the range of collectable samples.

\paragraph{Question construction.}
After obtaining the filtered dynamic charts, we construct candidate question-answer pairs for each chart. We use GPT-5 to generate 6 candidate questions for each chart, together with their corresponding answers, resulting in a candidate set of 4,038 question-answer pairs. The candidate set is then manually reviewed. The review contains two levels. The first level checks the quality of the question-answer pair itself. Annotators examine whether the question is clear, whether the answer is correct and unique, and whether the question can be supported by the chart content. The second level checks the interactivity of the question. Annotators examine whether the question can be answered from the initial static chart alone, and whether the answer is hidden in the HTML or underlying data but cannot be obtained through observable interaction. The former cannot reflect dynamic chart understanding, while the latter exceeds the scope of observable interaction. After review, candidate questions are either removed or revised. Questions with low quality, incorrect answers, or insufficient interaction requirements are removed. For charts with meaningful interactivity but imperfect question design, annotators revise the questions and answers so that the required evidence can be obtained through effective interaction. After candidate generation, manual review, and manual revision, we retain 1,440 high-quality question-answer pairs as the official evaluation samples of ChartAct.

\paragraph{Representative subset construction.}
In addition to the full evaluation set, we construct a representative subset of 300 samples for evaluating models with high inference cost. This subset is constructed before evaluating the target models, using the full-set results of 8 models with different scales and access types. We convert the success or failure of these models on each sample into a binary outcome matrix. Based on this matrix, we apply greedy selection followed by local swap search, and finally obtain a subset of 300 samples. This subset substantially reduces the evaluation cost of high-cost models while almost fully preserving the main model performance trends of the full set. The detailed algorithm is provided in the Appendix~\ref{sec:representative-subset}.

\subsection{Environment and Statistics}

\paragraph{Unified interaction environment.}
To fairly evaluate different models on dynamic chart tasks, ChartAct reduces the extra variation introduced by runtime platforms and interaction interfaces. We therefore adopt a unified interactive evaluation environment. This environment follows the desktop interaction framework of OSWorld \cite{DBLP:conf/nips/XieZCLZCHCSLLXZ24} and requires all models to complete tasks under the same setting. For each sample, a model observes the current page, performs interaction actions, and submits a final answer after obtaining sufficient evidence. This setting makes the comparison across models mainly reflect their dynamic chart understanding and interactive reasoning abilities. We adapt the original OSWorld runtime to dynamic chart question answering, including task prompts, trajectory turns, and history management. Detailed implementation choices are provided in the Appendix~\ref{sec:implementation-settings}.

\paragraph{Dashboard environment.}
To better match real use cases of dynamic charts, we further construct dashboard-style page environments. Dynamic charts in real webpages usually appear as data components and form analytical interfaces together with titles and other visual modules. Based on this observation, we use GPT-5.4 to generate diverse dashboard templates covering application scenarios such as finance, industry, law, and medicine. We then randomly embed the filtered dynamic charts into these templates, so that each chart appears in a concrete page context, which is shown in Figure~\ref{fig:overview}. This environment requires models to first locate the target chart in the page, and then interact with the chart to answer the question. Through this design, ChartAct evaluates dynamic chart understanding in concrete dashboard scenarios.

\paragraph{Statistics.}
Based on the above environment construction, ChartAct contains two paired evaluation environments. Dynamic Chart (DC) provides a clean dynamic chart environment, where the chart is directly displayed on the page. Dashboard Chart (DB) embeds the same dynamic chart into a dashboard page, and is used to evaluate the effect of page context and visual distractors on dynamic chart understanding. The two environments use the same questions and ground-truth answers, allowing paired comparison. ChartAct contains 1,440 question-answer samples, and each sample is instantiated in both DC and DB environments. The source and type distribution of the collected interactive charts is shown in Table~\ref{tab:chart-source-type}.

\subsection{Evaluation}
% ChartAct adopts an answer-driven evaluation setting. Each evaluation sample contains a question, a ground-truth answer, and a scoring rule. After observing and interacting with the environment, the model submits a final answer, and the evaluation determines task completion based on this answer. Since answers to dynamic chart questions may involve numbers, text, or natural-language expressions, we use an LLM judge to automatically assess model responses. The judge model determines whether the response is correct according to the question, the ground-truth answer, the scoring rule, and the model answer. To reduce the instability of a single judgment, we use multiple votes to obtain the final result. Each sample is finally marked as correct or incorrect. We use success rate as the main evaluation metric and report results by environment type and chart type. The judge prompt, voting setup, and scoring details are provided in the Appendix.

ChartAct adopts an answer-driven evaluation setting. Each evaluation sample contains a question and a ground-truth answer. After observing and interacting with the environment, the model submits a final answer, and the evaluation determines task completion based on this answer. Since answers to dynamic chart questions may involve numbers, text, or natural-language expressions, we use an LLM judge to automatically assess model responses. The judge model determines whether the response is correct according to the question, the ground-truth answer, and the model answer. To reduce the instability of a single judgment, we use multiple votes to obtain the final result. Each sample is finally marked as correct or incorrect. We use success rate as the main evaluation metric and report results by environment type and chart type. The judge prompt, voting setup, and scoring details are provided in the Appendix~\ref{sec:llm-judge-protocol}.

\begin{table*}[t]
\small
\centering
\setlength{\tabcolsep}{3.2pt}
\renewcommand{\arraystretch}{1.08}
\resizebox{\textwidth}{!}{%
\begin{tabular}{lrrrrrrrrrrr}
\toprule
& \multicolumn{4}{c}{Benchmark split} & \multicolumn{7}{c}{Chart-type score} \\
\cmidrule(lr){2-5}\cmidrule(lr){6-12}
Model & DC & DB & Avg. & Drop & Bar & Line & Pie & Scatter & Box & Heatmap & Other \\
\midrule
\rowcolor{gray!12}
\multicolumn{12}{l}{\emph{Proprietary models}} \\
Claude-Opus-4.7$^*$ & \textbf{85.3} & \textbf{83.7} & \textbf{84.5} & 1.7 & \textbf{88.7} & \textbf{84.7} & \textbf{91.0} & \textbf{79.2} & \textbf{86.2} & \textbf{75.0} & \textbf{77.8} \\
GPT-5.5$^*$ & 65.0 & 51.7 & 58.3 & 13.3 & 62.9 & 60.5 & 54.0 & 58.3 & 67.2 & 43.2 & 53.7 \\
Doubao-Seed-2.0-Pro & 55.6 & 39.4 & 47.5 & 16.1 & 53.5 & 47.3 & 41.7 & 49.1 & 53.3 & 45.1 & 37.6 \\
Gemini-3.1-Pro$^*$ & 61.0 & 31.0 & 46.0 & 30.0 & 47.6 & 47.6 & 50.0 & 49.0 & 46.6 & 40.9 & 29.6 \\
Qwen3.6-Plus & 54.7 & 23.8 & 39.2 & 30.8 & 46.5 & 36.3 & 36.2 & 39.6 & 45.7 & 43.2 & 24.4 \\
Qwen3-VL-Plus & 31.4 & 13.6 & 22.5 & 17.8 & 26.3 & 18.6 & 27.5 & 22.4 & 20.3 & 26.7 & 12.8 \\
\midrule
\rowcolor{gray!12}
\multicolumn{12}{l}{\emph{Open-source models}} \\
Kimi-K2.5 & \textbf{75.3} & \textbf{56.6} & \textbf{66.0} & 18.8 & \textbf{75.1} & \textbf{62.2} & \textbf{68.0} & \textbf{58.9} & \textbf{72.8} & \textbf{63.1} & \textbf{57.4} \\
Qwen3.5-122B-A10B & 51.9 & 31.9 & 41.9 & 20.0 & 52.7 & 41.8 & 34.1 & 41.5 & 45.3 & 44.7 & 26.4 \\
Qwen3.5-35B-A3B & 45.7 & 20.6 & 33.1 & 25.1 & 42.3 & 29.9 & 29.3 & 33.7 & 34.1 & 37.4 & 20.9 \\
GLM-4.6V & 38.7 & 23.0 & 30.8 & 15.7 & 36.8 & 25.3 & 29.8 & 32.0 & 33.3 & 41.3 & 19.0 \\
Gemma-4-31B-IT & 31.3 & 14.9 & 23.1 & 16.4 & 26.3 & 18.7 & 25.4 & 22.8 & 26.1 & 26.7 & 16.3 \\
\bottomrule
\end{tabular}%
}
\caption{Mixed-scope benchmark results with chart-type subscores. Full-benchmark models are evaluated on all 1,440 Dynamic Chart and 1,440 Dashboard Chart cases. Models marked with $^*$ are evaluated only on the representative subset of 300 cases per benchmark, and those subset scores are used as their benchmark estimates. Chart-type scores pool the two dataset variants. Scores are success rates in percent; bold values indicate the best success rate within each access block.}
\label{tab:subset-results}

\end{table*}

\begin{figure*}[ht]
    \centering
    \includegraphics[width=0.9\linewidth]{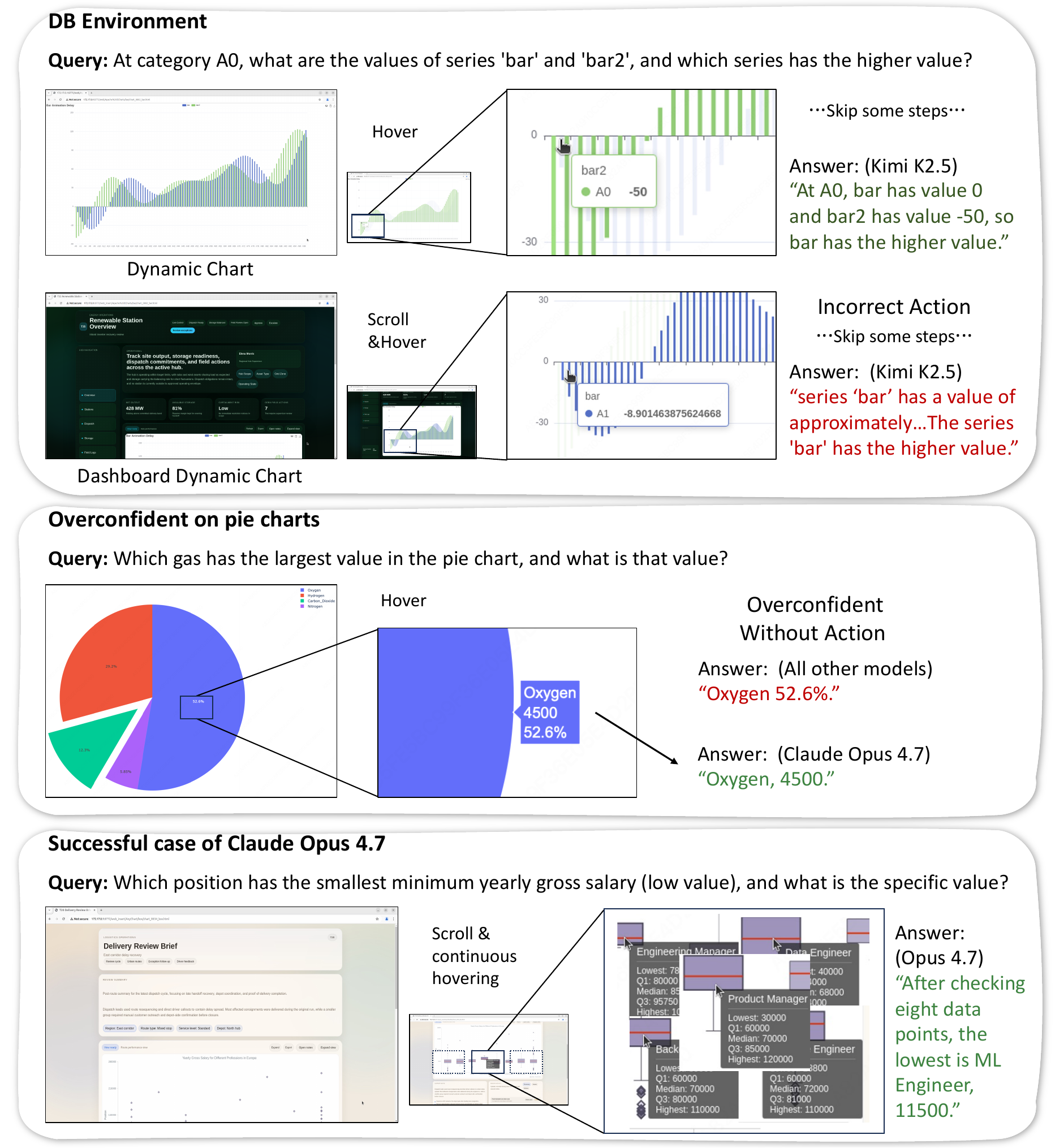}
    \caption{Case studies of model behaviors in ChartAct. The examples show dashboard context causing incorrect interaction, overconfident answering on pie charts without sufficient interaction, and Claude-Opus-4.7 solving a boxplot question through systematic scrolling and continuous hovering.}

    \label{fig:case}
\end{figure*}

% For models evaluated on the representative subsets, the reported DC and DB scores are subset-based estimates of their full-benchmark performance. Since the 300-case subsets for Dynamic Chart and Dashboard Chart are selected independently, the corresponding Drop value should be interpreted as an estimated robustness gap rather than a strictly paired case-level comparison. The strictly paired DC--DB comparison is preserved for the full benchmark, where the two variants share the same questions and ground-truth answers.
\section{Experiments}

\subsection{Experimental Setup}

We evaluate current advanced multimodal models and GUI agents on ChartAct. All models use a unified interaction framework, with the same page observations, action interface, task prompts, and answer submission format. ChartAct contains two environments: Dynamic Chart (DC) and Dashboard Chart (DB). DC directly presents the dynamic chart, while DB embeds the same dynamic chart into a dashboard page. The two environments share the same questions and answers, so the performance gap between DC and DB for the same model reflects the additional effect of the dashboard context. For models with high inference cost, we evaluate them on 300 representative samples. This subset has been verified to closely preserve the overall trend, so the corresponding results are used as benchmark estimates and marked with a star in the table. All answers are judged by an LLM judge according to the ground-truth answer and scoring rule. The detailed evaluation criteria are provided in the Appendix.

\subsection{Main observations}

Table~\ref{tab:subset-results} provides an overall view of model performance on ChartAct. The results reveal three clear patterns.

\paragraph{ChartAct remains challenging.}
    Claude-Opus-4.7 achieves the highest average success rate of 84.5\%, but this performance does not appear broadly across other models. Kimi-K2.5 drops to 66.0\%, GPT-5.5 reaches 58.3\%, and Doubao-Seed-2.0-Pro and Gemini-3.1-Pro further drop to 47.5\% and 46.0\%. The clear gap between the best result and the remaining models shows that most models still struggle to understand charts that require interaction.

\paragraph{Proprietary models show a higher overall ceiling.}
    Claude-Opus-4.7 ranks first, and GPT-5.5, Doubao-Seed-2.0-Pro, and Gemini-3.1-Pro also fall in the higher range. Among open-source models, the large-scale Kimi-K2.5 performs strongly. Excluding Kimi-K2.5, the best open-source model is Qwen3.5-122B-A10B, with an average success rate of 41.9\%. This distribution shows that current dynamic chart understanding ability is more concentrated in stronger models, while open-source models still have substantial room for improvement.

% \paragraph{DB brings a performance drop for all models.}
%     Every model in the table obtains a lower success rate on DB than on DC. Claude-Opus-4.7 drops by only 1.7 percentage points, indicating that the strongest model already has better ability to handle page context. Gemini-3.1-Pro and Qwen3.6-Plus drop by 30.0 and 30.8 percentage points, indicating that many models cannot directly transfer their ability from clean chart pages to dashboard pages. Since DC and DB use the same charts and questions, this gap directly corresponds to the difficulty introduced by the complex page environment.
\paragraph{DB brings a performance drop for all models.}
Every model in the table obtains a lower success rate on DB than on DC, showing that dashboard-style page context consistently increases the difficulty of dynamic chart understanding. Claude-Opus-4.7 drops by only 1.7 percentage points, indicating that the strongest model has a better ability to handle page context. Gemini-3.1-Pro and Qwen3.6-Plus drop by 30.0 and 30.8 percentage points, indicating that many models cannot directly transfer their ability from clean chart pages to dashboard pages. For models evaluated on the full benchmark, DC and DB use the same charts and questions, so this gap directly reflects the difficulty introduced by the complex page environment. For subset-only models marked with $\ast$, the DC and DB scores are computed on independently selected representative subsets; therefore, their Drop values should be interpreted as numerical estimates of performance degradation rather than strict case-level paired comparisons between each chart and its dashboard counterpart.

\subsection{Analysis}

After splitting the results by chart type, Other obtains the lowest score. This result is intuitive, because Other covers more diverse chart structures, more irregular interaction targets, and a larger search space. Since Other is not a single chart type, we treat it only as a supplementary observation. The main observations are as follows.

\paragraph{Line has the largest drop from DC to DB.}
Line is not the lowest-performing type in the clean environment, but it drops the most after entering the dashboard environment. This change shows that the difficulty of Line is mainly amplified by page context. Many Line questions require models to trigger the correct tooltip along continuous x-axis positions, while also distinguishing adjacent lines or adjacent data points. Scrolling, resizing, and surrounding components in dashboards increase the complexity of the environment and therefore substantially increase the difficulty of understanding.

\paragraph{Pie has the smallest drop, but Pie itself is not simple.}
Pie already performs relatively low in DC, indicating that its main errors come from inside the chart. Pie lacks regular coordinate axes, and models often ignore interaction and directly estimate or guess the answer. Claude-Opus-4.7 reaches 91.0\% on Pie, showing that this task can be solved through effective interaction. Weaker models perform much lower on Pie, further supporting that their errors often come from direct estimation or guessing. The additional DB drop for Pie is small because its data density and environmental complexity are already low.

\paragraph{Chart types do not form a fixed order.}
Claude-Opus-4.7 achieves its highest score on Pie, GPT-5.5 achieves its highest score on Box, and Kimi-K2.5 and Qwen3.5-122B-A10B achieve their highest scores on Bar. This phenomenon shows that the difficulty of dynamic charts cannot be explained only by chart type. Environmental factors such as the position of interaction regions and whether evidence is hidden can change model performance within the same chart type.

% \subsection{Failure Attribution}

% From the error trajectories, model failures mainly come from three types.

% \paragraph{Evidence is not extracted}. Models answer from the initial screenshot without triggering chart state changes. Such answers often look plausible, but fail directly on questions that require exact values.

% \paragraph{Interaction position shifts}. The model has determined that it needs to operate on the chart, but the mouse lands on an adjacent data point, an adjacent sector, or an invalid region. The wrong landing point leads to wrong feedback or no usable feedback.

% \paragraph{Page object is misidentified}. In DB, titles, metric cards, and other visualization components appear around the target chart. Some models treat these components as the target chart and continue operating on the wrong object, causing the subsequent trajectory to deviate from the task.
\subsection{Failure Attribution}

Based on the observed error trajectories, model failures can be mainly attributed to three categories.

\paragraph{Failure to extract evidence.}
Some models answer the question directly from the initial screenshot without triggering the interactive state changes of the chart. Although such answers may appear plausible, they often fail on questions that require precise values revealed only through interaction.

\paragraph{Mis-localized interactions.}
The model correctly infers that chart interaction is required, but the cursor is placed on a neighboring data point, an adjacent sector, or an invalid region. This spatial misalignment produces incorrect feedback or no usable feedback, which then leads to an erroneous answer.

\paragraph{Target object misidentification.}
In DB, titles, metric cards, and other visualization components may appear near the target chart. Some models mistakenly regard these surrounding components as the target chart and continue interacting with the wrong object, causing the subsequent trajectory to deviate from the intended task.

% \paragraph{Failure to recover from ineffective actions.}
% Models sometimes recognize that additional evidence is needed, but fail to revise their strategy when the current interaction does not produce useful feedback.
% They may repeatedly hover, click, or scroll in nearly the same region, or submit a final answer despite explicitly uncertain or incomplete evidence. This failure differs from a single mislocalized interaction: the key problem is the lack of
% self-correction after observing that the previous action was ineffective.

\subsection{Case Study}

We conduct case studies in Figure~\ref{fig:case}. The figure shows three representative trajectories.

\paragraph{Dashboard context changes the interaction trajectory.}
The first case compares the same question in DC and DB. In DC, the model directly locates the bar chart and obtains the two series values at A0 by hovering. In DB, the denser page context shifts the interaction to A1, producing a wrong observation and then a wrong answer. This case shows how DB changes the interaction path.

\paragraph{Pie charts expose overconfident answering.}
The second case shows an overconfident error on a pie chart. Most models answer with the visible percentage, Oxygen 52.6\%, without observing the true value. Claude-Opus-4.7 hovers over the sector and obtains the correct value, Oxygen 4500. This case shows that pie charts can look simple while still requiring interaction.

\paragraph{Strong models conduct systematic evidence search.}
The third case shows a successful trajectory of Claude-Opus-4.7. The model scrolls to the target boxplot and continuously hovers over multiple boxes. By comparing the observed low values, it identifies ML Engineer with 11500. This case shows the benefit of continuous interaction and evidence comparison.

% \begin{figure}[t]
%     \centering
%     \includegraphics[width=\linewidth]{case_pic/case1-1.png}
%     \caption{Located successfully in DC.}
%     \label{fig:subset-pipeline}
% \end{figure}

% \begin{figure}[t]
%     \centering
%     \includegraphics[width=\linewidth]{case_pic/case1-2.png}
%     \caption{Locating failed in DB.}
%     \label{fig:subset-pipeline}
% \end{figure}

\section{Conclusion}

We propose ChartAct, a benchmark for dynamic chart understanding. ChartAct requires models to answer chart questions through actions, observations, and reasoning. Experimental results show that current multimodal models and GUI agents still face clear challenges in dynamic chart understanding. Models often fail in interactive evidence acquisition and fine-grained operations. These results show that dynamic charts introduce evaluation requirements beyond static chart understanding. ChartAct provides a new testbed for studying chart understanding in real interactive environments.

\section*{Limitations}

ChartAct mainly focuses on dynamic charts from real websites and evaluates them in controllable interactive environments. Although the data covers diverse chart sources and dashboard scenarios, it cannot include all chart libraries and real dashboard layouts. The current evaluation mainly relies on final answers, which measures task completion but cannot fully characterize the quality of intermediate interactions. Future work can further expand chart sources, interaction types, and process-level evaluation.

% \newpage
\bibliography{custom}

\appendix

% Float placement tuning for the appendix. This only affects float placement.
\setcounter{topnumber}{5}
\setcounter{bottomnumber}{5}
\setcounter{totalnumber}{8}
\renewcommand{\topfraction}{0.95}
\renewcommand{\bottomfraction}{0.90}
\renewcommand{\textfraction}{0.05}
\renewcommand{\floatpagefraction}{0.75}
\renewcommand{\dbltopfraction}{0.95}
\renewcommand{\dblfloatpagefraction}{0.75}

% \section{Example Appendix}
% \label{sec:appendix}
% 附录这块有三个需要写
% 1.子集生成算法流程
% 2.OSworld关于运行的框架的调整，参数调整和提示词
% 3.模型判卷子的流程

\section{Representative Subset Construction}
\label{sec:representative-subset}

This appendix describes how we construct the 300-case representative subsets used
for high-cost model evaluation. ChartAct contains two benchmark variants,
Dynamic Chart (DC) and Dashboard Chart (DB). Each variant contains 1,440 cases.
Evaluating every high-cost model on all $2 \times 1440$ cases is expensive, so
we construct one 300-case subset for DC and one 300-case subset for DB. The two
subsets are selected independently, but they are generated using the same
selection algorithm and the same objective design. Importantly, the target
high-cost models are not involved in subset construction. The subsets are
constructed only from the full-set results of eight lower-cost pilot models.Specifically, the eight pilot models are Doubao-Seed-2.0-Pro, Kimi-K2.5, Qwen3.6-Plus, Qwen3-VL-Plus, Gemma-4-31B-IT, GLM-4.6V, Qwen3.5-122B-A10B, and Qwen3.5-35B-A3B. The subset-only high-cost models are excluded from this construction process to avoid using their evaluation outcomes during subset selection.

\subsection{Pilot Outcome Matrix}

For each benchmark variant, we first collect the full-set binary results of the
eight pilot models. Let $D=\{x_i\}_{i=1}^{N}$ denote one full benchmark variant,
where $N=1440$, and let $M=8$ be the number of pilot models. We construct a
binary outcome matrix
\[
Y \in \{0,1\}^{N \times M},
\]
where $Y_{ij}=1$ means that pilot model $j$ correctly solves case $x_i$, and
$Y_{ij}=0$ otherwise. Each row of $Y$ therefore summarizes how the eight pilot
models behave on one case.

From this matrix, each case receives two types of calibration information. The
first is its empirical difficulty,
\[
d_i=\sum_{j=1}^{M}Y_{ij},
\]
which is the number of pilot models that solve the case. A case with $d_i=8$ is
easy for all pilot models, while a case with $d_i=0$ is difficult for all pilot
models. Intermediate values indicate cases that separate the pilot models. The
second is the chart type of the case, such as line, bar, pie, scatter, heatmap,
box, or other. These features allow the subset selection process to preserve
both performance behavior and content coverage.

\subsection{Representativeness Objective}

The goal is to select a subset $S \subset D$ with $|S|=300$ such that $S$ behaves
as similarly as possible to the full benchmark $D$. We define a weighted
representativeness objective with four terms:
\[
\begin{array}{rl}
\mathcal{L}(S) = &
5.0\,\Delta_{\mathrm{acc}}(S,D)
+ 2.0\,\Delta_{\mathrm{gap}}(S,D) \\
&+ 2.0\,\Delta_{\mathrm{diff}}(S,D)
+ 1.2\,\Delta_{\mathrm{type}}(S,D).
\end{array}
\]
Lower values of $\mathcal{L}(S)$ indicate a more representative subset.

\paragraph{Model accuracy deviation.}
For pilot model $j$, let
\[
a_j(S)=\frac{1}{|S|}\sum_{x_i \in S}Y_{ij}
\]
be its accuracy on subset $S$, and define $a_j(D)$ analogously on the full
benchmark. The model-accuracy deviation is
\[
\Delta_{\mathrm{acc}}(S,D)
=\frac{1}{M}\sum_{j=1}^{M}\left|a_j(S)-a_j(D)\right|.
\]
This term makes the selected subset preserve the absolute performance level of
each pilot model.

\paragraph{Pairwise model-gap deviation.}
Matching only individual accuracies is not sufficient; the subset should also
preserve the relative gaps between models. For each pair of pilot models
$(u,v)$, define the accuracy gap on subset $S$ as
\[
g_{uv}(S)=a_u(S)-a_v(S).
\]
The pairwise-gap deviation is
\[
\Delta_{\mathrm{gap}}(S,D)
=\frac{1}{\binom{M}{2}}\sum_{1 \leq u < v \leq M}
\left|g_{uv}(S)-g_{uv}(D)\right|.
\]
This term encourages the subset to preserve the model ordering and the relative
separation between pilot models.

\paragraph{Difficulty-distribution deviation.}
Let $p_b^{\mathrm{diff}}(S)$ be the proportion of cases in $S$ whose difficulty
is $b$, where $b \in \{0,1,\ldots,8\}$. The difficulty-distribution deviation is
\[
\Delta_{\mathrm{diff}}(S,D)
=\sum_{b=0}^{8}
\left|p_b^{\mathrm{diff}}(S)-p_b^{\mathrm{diff}}(D)\right|.
\]
This term prevents the subset from becoming systematically easier or harder than
the full benchmark.

\paragraph{Chart-type distribution deviation.}
Let $p_t^{\mathrm{type}}(S)$ be the proportion of cases in $S$ with chart type
$t$. The chart-type distribution deviation is
\[
\Delta_{\mathrm{type}}(S,D)
=\sum_{t}
\left|p_t^{\mathrm{type}}(S)-p_t^{\mathrm{type}}(D)\right|.
\]
This term preserves the content composition of the full benchmark across major
chart categories.

\subsection{Greedy Construction}

After defining the objective, we search for a low-score subset using greedy
construction followed by local swap search. For each benchmark variant, the
algorithm starts with an empty selected set $S=\emptyset$ and an available pool
$U=D$. It then repeatedly adds one case until $|S|=300$.

At greedy step $k$, where $0 \leq k < 300$, the algorithm evaluates every
currently unselected candidate case $x \in U$. For each candidate, it
temporarily forms
\[
S_x = S \cup \{x\}
\]
and computes the objective value $\mathcal{L}(S_x)$. All candidate cases are
then ranked by this score in ascending order. The lowest-score candidates are
the cases that make the current subset most similar to the full benchmark after
being added.

To avoid a brittle deterministic path, we do not always select only the single
best candidate. Instead, we take the top $K_{\mathrm{top}}=8$ candidates and
sample one candidate from them using rank-based probabilities. If a candidate is
ranked $r$ among the top candidates, its sampling weight is
\[
w_r=\frac{1}{r}.
\]
Thus, the best candidate has the largest probability, but lower-ranked
candidates among the top eight can still be selected. This small amount of
randomness reduces the risk that an early greedy choice traps the search in an
inferior local solution. We use a fixed random seed for reproducibility. The DC
subset uses seed 20260521, and the DB subset uses seed 20261521.

The greedy stage can be summarized as follows:
\begin{enumerate}
    \item Initialize $S=\emptyset$.
    \item For every unselected case $x$, compute $\mathcal{L}(S\cup\{x\})$.
    \item Rank all candidates by this objective value.
    \item Sample one case from the top eight candidates with probability
    proportional to $1/r$, where $r$ is the candidate rank.
    \item Add the sampled case to $S$.
    \item Repeat until $|S|=300$.
\end{enumerate}

\subsection{Local Swap Search}

The greedy stage constructs a strong initial subset, but greedy selection only
optimizes the immediate next addition. A case selected early may become
suboptimal after many later additions. To further improve the subset, we apply a
local swap search after the greedy subset reaches 300 cases.

Let $S$ be the current 300-case subset and let $D \setminus S$ be the set of
unselected cases. At each local-search iteration, the algorithm randomly selects
one case $x_{\mathrm{out}} \in S$ and one case $x_{\mathrm{in}} \in D \setminus
S$. It proposes a swapped subset
\[
S'=(S \setminus \{x_{\mathrm{out}}\}) \cup \{x_{\mathrm{in}}\}.
\]
The swap is accepted only if it strictly reduces the representativeness
objective:
\[
\mathcal{L}(S') < \mathcal{L}(S).
\]
If the objective does not improve, the swap is rejected and the algorithm keeps
the original subset. Therefore, the local-search stage monotonically decreases
the objective value over accepted moves.

In our implementation, the local search runs for at most 25,000 swap trials for
each benchmark variant. We also use an early stopping rule: if the search
encounters 5,000 consecutive non-improving swap proposals, it stops early. The
output of this stage is a locally optimized 300-case subset.

\subsection{Final Subsets and Validation}

We run the above procedure independently for DC and DB. This produces two
300-case subsets, one for each benchmark variant. The resulting subsets reduce
the number of evaluated cases from $2 \times 1440$ to $2 \times 300$, reducing
the evaluation load by approximately 79.2\%.

We validate the selected subsets by comparing pilot-model performance on the
300-case subsets against performance on the corresponding full 1,440-case
benchmarks. For DC, the mean absolute pilot-model accuracy error is 0.000590 and
the maximum absolute accuracy error is 0.002083. For DB, the mean absolute
pilot-model accuracy error is 0.001267 and the maximum absolute accuracy error
is 0.002222. In both benchmark variants, the pilot-model ranking is preserved
exactly, with zero rank inversions. These results indicate that the selected
subsets closely preserve the evaluation behavior of the full benchmark while
substantially reducing evaluation cost.

% 中文解释版本，仅供作者理解，不参与正式英文论文编译。
% 本附录描述的是我们如何从两个完整测试集各 1440 个 case 中，分别选出 300 个代表性 case。
% 这两个小集合分别对应 Dynamic Chart 和 Dashboard Chart。两个集合独立生成，但使用同一套算法。
% 目标不是随机抽样，也不是只挑难题，而是让 300 个 case 在模型表现和内容组成上尽量像完整 1440 个 case。
%
% 第一步是构建 pilot outcome matrix。我们使用 8 个低成本 pilot model 的全量测试结果。
% 对每个 case，记录这 8 个模型是否做对，形成一个 0/1 向量。例如 [1,0,1,1,0,0,1,0]。
% 这个向量可以反映该 case 对模型的难度和区分作用。
%
% 第二步是定义目标函数。我们只保留四个核心偏差项：
% 1. 模型准确率偏差，权重 5.0；
% 2. 模型两两差距偏差，权重 2.0；
% 3. 难度分布偏差，权重 2.0；
% 4. chart type 分布偏差，权重 1.2。
% 目标函数越小，说明当前子集越接近完整集合。
%
% 第三步是贪心构造。算法从空集合开始，每一步尝试把每一个未选 case 加入当前集合，
% 计算加入后目标函数会变成多少。然后按目标函数从小到大排序。
% 不是永远选择第一名，而是从前 8 名中按排名概率抽取一个 case。
% 第 1 名权重为 1，第 2 名权重为 1/2，第 3 名权重为 1/3，以此类推。
% 这样既保留贪心方向，又避免早期选择过于死板。
%
% 第四步是局部交换搜索。贪心选满 300 个之后，算法随机从已选集合中拿出一个 case，
% 再从未选集合中拿进一个 case，形成一个新的候选集合。如果交换后目标函数更低，就接受交换；
% 如果没有变好，就撤销交换。每个测试集最多尝试 25000 次交换。如果连续 5000 次没有改进，就提前停止。
%
% 最终得到两个 300-case 小集合。验证结果显示，Dynamic Chart 子集的平均 pilot-model 准确率误差为 0.000590，
% Dashboard Chart 子集为 0.001267，并且两个子集都保持了 pilot model 的完整排序，rank inversion 为 0。

\section{Implementation Details and Experimental Settings}
\label{sec:implementation-settings}

This appendix summarizes the implementation changes we made on top of the
original OSWorld-style evaluation framework and the hyperparameters used in our
experiments. The goal of these changes is to adapt the desktop automation
runtime to answer-driven dynamic chart understanding while keeping the
interaction interface as consistent as possible across models.

\subsection{Incremental Changes over OSWorld}

Our runtime follows the basic OSWorld design: each task is executed in a
virtualized Ubuntu desktop, the agent observes the current screen, emits GUI
actions, and the environment executes those actions before returning the next
observation. We make several task-specific modifications for ChartAct.

\paragraph{Chart-specific prompting.}
We replace the generic desktop-task prompt with a chart-focused prompt in
\texttt{mm\_agents/prompts.py}. The main prompt explicitly defines the task as
interacting with charts already displayed in a browser, collecting information
from chart states, and answering chart-related questions.  In particular, models are not allowed to use developer tools,
inspect page source, call screenshot APIs, or rely on hidden webpage metadata. Figure~\ref{fig:prompt-excerpt} shows the core excerpt of the modified interaction prompt used in the main framework.

\begin{figure}[!t]
\centering
\setlength{\fboxsep}{4pt}
\fbox{%
\begin{minipage}{0.92\columnwidth}
\footnotesize
\textbf{Core excerpt of the ChartAct interaction prompt}

\vspace{0.4em}
\ttfamily
Task: interact with charts already displayed on a webpage to collect
information and answer chart-related questions.

\vspace{0.4em}
Use only normal browser interactions. Do not use developer mode or access the
webpage source code.

\vspace{0.4em}
Case 1: Perform an Interaction. Return Python code in exactly one code block.
Do not use pyautogui.locateCenterOnScreen or pyautogui.screenshot().

\vspace{0.4em}
Case 2: Answer the Question. Return FINAL\_JSON:
\{"Answer":"you fill your answer here"\}.

\vspace{0.4em}
If CURRENT\_STEP equals MAX\_STEPS, do not output Python code; output the final
answer instead.
\end{minipage}}
\caption{Core excerpt of the modified interaction prompt used in the main
framework.}
\label{fig:prompt-excerpt}
\end{figure}

\paragraph{Answer-driven termination.}
The original OSWorld tasks are often evaluated through the final desktop state.
In ChartAct, each task is a question-answering problem. We therefore add a
machine-readable final-answer channel:
\[
\texttt{FINAL\_JSON: \{"Answer":"..."\}}.
\]
During evaluation, \texttt{lib\_run\_single.py} extracts this field, writes it
to \texttt{final\_answer.txt}, and stores it in \texttt{env.agent\_answer}. The
environment evaluator then grades this submitted answer rather than treating a
generic \texttt{DONE} action as sufficient.

\paragraph{Step-aware interaction control.}
We modify the agent wrapper in \texttt{mm\_agents/agent.py} to pass the current
step index into the prompt. When the maximum step budget is reached, the prompt
explicitly forbids further Python actions and requires the model to submit its
best final answer. This prevents models from spending the final turn on an
action for which no follow-up observation will be available.

\paragraph{Unified model routing.}
We extend the model-calling code in \texttt{mm\_agents/agent.py} so that the main
evaluation framework routes different model families through a common
OpenAI-compatible chat-completion interface whenever possible. The routing layer
normalizes provider-specific base URLs, API keys, token fields, and
reasoning/thinking options, while preserving the same visible task prompt,
screenshot input, action grammar, history window, and final-answer format for
the main comparison.

\paragraph{Execution safety and logging.}
The environment executes interaction actions through the \texttt{pyautogui}
action space. Before execution, generated Python commands are sanitized to
remove non-GUI or dangerous statements. For each task, the runner stores the
initial screenshot, per-step screenshots, structured trajectory logs, the final
answer, a screen recording, and the scalar result. These artifacts make the
evaluation auditable and allow later error analysis without rerunning the
interaction.

\begin{table*}[!t]
\centering
\small
\setlength{\tabcolsep}{5pt}
\renewcommand{\arraystretch}{1.08}
\begin{tabular}{@{}p{0.22\textwidth}p{0.72\textwidth}@{}}
\toprule
\textbf{Parameter} & \textbf{Setting} \\
\midrule
Main runner & \texttt{run\_multienv.py} \\
Main agent wrapper & \texttt{PromptAgent} in \texttt{mm\_agents/agent.py} \\
Environment provider & Docker-based OSWorld desktop environment \\
Operating system and browser & Ubuntu desktop with Google Chrome opened to the target chart page \\
Screen resolution & \(1920 \times 1080\) \\
Observation type & Screenshot-only input in the reported main setting \\
Action space & \texttt{pyautogui} desktop-control code \\
Maximum interaction budget & 15 effective environment steps per task \\
Agent-side history window & Last 7 trajectory turns (\texttt{max\_trajectory\_length=7}) \\
Post-action wait & 1.0 second after each executed action \\
Sampling temperature & 0.3 \\
Top-\(p\) & 0.9 \\
Maximum output tokens & 3,000 tokens per model call \\
Final-answer format & \texttt{FINAL\_JSON: \{"Answer":"..."\}} \\
Result directory format & \texttt{results/pyautogui/screenshot/<model>/...} \\
Evaluation metric & Answer-based \texttt{llm\_judge} with configurable majority voting \\
Parallel environments & Varied only for throughput according to API and machine capacity; this does not change the per-task protocol \\
\bottomrule
\end{tabular}
\caption{Shared experimental settings used by the main fair-comparison
framework. The number of parallel environments is varied only for throughput and
does not change the per-task interaction protocol.}
\label{tab:shared-settings}
\end{table*}

\subsection{Shared Evaluation Parameters}

Table~\ref{tab:shared-settings} lists the parameters kept fixed in the main
fair-comparison framework. We do not manually set the internal context length of
each proprietary model. Instead, we control the agent-side context window by
keeping at most the latest seven trajectory turns in the prompt. The maximum
model output length is fixed to 3,000 tokens for all reported runs.

The dataset selector is changed according to the evaluated benchmark split. For
full-set experiments, the metadata files are \texttt{chartall.json} for Dynamic
Chart and \texttt{webchartall.json} for Dashboard Chart. For cost-controlled
evaluation, the same framework is run on the representative subset metadata
files, such as \texttt{chart\_300.json} and \texttt{webchart\_300.json}. These
metadata choices select the task pool; they do not alter the interaction
interface or model-facing prompt.

\subsection{Model-Specific Interface Settings}

The experiments involve eight reported model entries, all evaluated with the
same main interaction framework. Table~\ref{tab:model-settings} summarizes the
model-specific interface settings used by the unified routing layer.

\begin{table*}[!t]
\centering
\small
\setlength{\tabcolsep}{5pt}
\renewcommand{\arraystretch}{1.06}
\begin{tabular}{@{}p{0.30\textwidth}p{0.64\textwidth}@{}}
\toprule
\textbf{Model entry} & \textbf{Model-specific interface setting} \\
\midrule
\multicolumn{2}{@{}l}{\emph{Full benchmark: \(2 \times 1440\) cases}} \\
\texttt{doubao-seed-2-0-pro-260215} & Seed/Doubao OpenAI-compatible route; reasoning depth set to \texttt{high}. \\
\texttt{kimi-k2.5} & Kimi OpenAI-compatible route; Kimi thinking switch set to \texttt{disabled}. \\
\texttt{qwen3.6-plus} & Qwen OpenAI-compatible route; Qwen thinking-token budget set to 1000 in the recorded run. \\
\texttt{qwen3-vl-plus} & Qwen OpenAI-compatible route; DashScope thinking enabled with thinking budget 2100. \\
\texttt{gemma-4-31b-it} & Open-weight model served through the unified OpenAI-compatible route; no extra reasoning switch. \\
\texttt{glm-4.6v} & Open-weight GLM route; response postprocessing removes reasoning tags before action parsing. \\
\texttt{qwen3.5-122b-a10b} & Open-weight Qwen route; Qwen thinking-token budget set to 2100. \\
\texttt{qwen3.5-35b-a3b} & Open-weight Qwen route; Qwen thinking-token budget set to 1500 in the recorded run. \\
\midrule
\multicolumn{2}{@{}l}{\emph{Representative subset: \(2 \times 300\) cases}} \\
\texttt{claude-opus-4-7} & Claude OpenAI-compatible route; unsupported sampling parameters are removed by the routing layer. \\
\texttt{gpt-5.5} & GPT OpenAI-compatible route; GPT-5 payload uses \texttt{max\_completion\_tokens} for the 3,000-token output cap. \\
\texttt{gemini-3.1-pro-preview} & Gemini OpenAI-compatible route; Gemini thinking depth set to \texttt{low} through provider-specific thinking configuration. \\
\bottomrule
\end{tabular}
\caption{Reported model entries and model-specific interface settings. Full and
subset entries use the same main wrapper, prompt, action space, history window,
step budget, answer format, and grading protocol; subset entries are evaluated
on the representative 300-case split for each benchmark variant.}
\label{tab:model-settings}
\end{table*}

\section{LLM Judge Grading Protocol}
\label{sec:llm-judge-protocol}

ChartAct uses answer-driven grading. Each task stores a question, a reference
answer, and a rubric in the task JSON. After the agent submits
\texttt{FINAL\_JSON}, the evaluator retrieves the cached \texttt{Answer} field
through the \texttt{agent\_answer} getter and passes it to
\texttt{llm\_judge}. Missing or empty answers are directly assigned score
\(0.0\).

\subsection{Judge Model and Voting Setup}

We use \texttt{deepseek-v4-pro} as the LLM judge. The judge is called through the
DeepSeek API endpoint \texttt{https://api.deepseek.com}. The judge configuration
uses three independent votes and an acceptance threshold of two votes. In other
words, a task is marked correct only if at least two of the three judge calls
return a positive decision. The judge decoding temperature is set to 0.0, the
maximum output length is 1,024 tokens, the timeout is 45 seconds, and each vote
allows one retry with a two-second retry backoff. We set the judge reasoning
effort to \texttt{low} and enable the provider-side thinking mode. API keys are
loaded from the local judge configuration file and are not included in the
paper.

% \begin{table}[!t]
% \centering
% \footnotesize
% \begin{tabular}{@{}ll@{}}
% \toprule
% \textbf{Judge setting} & \textbf{Value} \\
% \midrule
% Judge model & \texttt{deepseek-v4-pro} \\
% Base URL & \texttt{https://api.deepseek.com} \\
% Votes & 3 \\
% Acceptance threshold & 2 passing votes \\
% Temperature & 0.0 \\
% Maximum output tokens & 1,024 \\
% Reasoning effort & \texttt{low} \\
% Thinking mode & enabled \\
% Timeout & 45 seconds \\
% Retries & 1 retry per vote \\
% Retry backoff & 2 seconds \\
% \bottomrule
% \end{tabular}
% \caption{LLM judge configuration used for answer grading.}
% \label{tab:judge-config}
% \end{table}

\subsection{Judge Prompt}

For each vote, the judge receives a system message and a user message. The
system message enforces binary scoring, while the user message provides the task
question, the ground-truth answer, the agent answer, and the task-specific
rubric. The complete prompt template is shown in
Figure~\ref{fig:judge-prompt}. The placeholders are filled from the task JSON
and the submitted agent answer.

% \begin{figure*}[t]
% \centering
% \setlength{\fboxsep}{5pt}
% \fbox{%
% \begin{minipage}{0.94\textwidth}
% \footnotesize
% \textbf{System message}

% \vspace{0.3em}
% \ttfamily
% You are a strict judge. Output ONLY 1.0 or 0.0.

% \vspace{0.8em}
% \normalfont\footnotesize
% \textbf{User message}

% \vspace{0.3em}
% \ttfamily
% [Task Question]\\
% \{question\}

% \vspace{0.5em}
% [Ground Truth / Expected]\\
% \{expected\}

% \vspace{0.5em}
% [Agent's Actual Answer]\\
% \{extracted\_answer\}

% \vspace{0.5em}
% [Rubric]\\
% \{rubric\}

% \vspace{0.5em}
% Based on the rubric, is the Agent's answer correct?\\
% Output 1.0 for Yes, 0.0 for No.\\
% When you compare answers, focus on the actual correctness of the answers rather than wording or language.\\
% If the meaning of the answer is the same as the standard answer, give full marks.\\
% For numerical questions, rounding is allowed when the precision remains reasonable.
% \end{minipage}}
% \caption{Complete prompt template used by the LLM judge.}
% \label{fig:judge-prompt}
% \end{figure*}

\begin{figure*}[!t]
\centering
\setlength{\fboxsep}{5pt}
\fbox{%
\begin{minipage}{0.94\textwidth}
\footnotesize
\textbf{System message}

\vspace{0.3em}
\ttfamily
You are a strict judge. Output ONLY 1.0 or 0.0.

\vspace{0.8em}
\normalfont\footnotesize
\textbf{User message}

\vspace{0.3em}
\ttfamily
[Task Question]\\
\{question\}

\vspace{0.5em}
[Ground Truth / Expected]\\
\{expected\}

\vspace{0.5em}
[Agent's Actual Answer]\\
\{extracted\_answer\}

\vspace{0.5em}
Based on the task question and the ground-truth answer, is the Agent's answer correct?\\
Output 1.0 for Yes, 0.0 for No.\\
When you compare answers, focus on the actual correctness of the answers rather than wording or language.\\
If the meaning of the answer is the same as the standard answer, give full marks.\\
For numerical questions, rounding is allowed when the precision remains reasonable.
\end{minipage}}
\caption{Effective prompt template used by the LLM judge for ChartAct grading.}
\label{fig:judge-prompt}
\end{figure*}

\subsection{Score Aggregation and Regrading}

% The judge output parser extracts the first valid binary score, either
% \texttt{1.0} or \texttt{0.0}. A vote is considered passing if the parsed score is
% at least 0.5. Let \(v\) be the number of passing votes among the three judge
% calls. The final task score is
% \[
% \mathrm{score} =
% \begin{cases}
% 1.0, & v \geq 2,\\
% 0.0, & v < 2.
% \end{cases}
% \]
% The benchmark success rate is the mean of these binary task scores.
The judge output parser extracts the first valid binary score, either
\texttt{1.0} or \texttt{0.0}. Since the judge output is binary, a parsed score
of \texttt{1.0} is counted as a passing vote, while a parsed score of
\texttt{0.0} is counted as a failing vote. Let \(v\) be the number of passing
votes among the three judge calls. The final task score is
\[
\mathrm{score} =
\begin{cases}
1.0, & v \geq 2,\\
0.0, & v < 2.
\end{cases}
\]
The benchmark success rate is the mean of these binary task scores.

We also design an offline grading script for post-hoc evaluation after the
interaction trajectories have been completed. The script scans existing result
directories, reloads each task configuration, reads the saved
\texttt{final\_answer.txt}, reruns the current \texttt{llm\_judge}, and
optionally updates the stored grading result. Thus, if the judge model, voting
configuration, or rubric is updated, we can recompute scores from saved answers
without rerunning the expensive GUI interaction trajectories.

\end{document}